\documentclass{article}

\PassOptionsToPackage{numbers, compress}{natbib}

    \usepackage[final]{timeseries_workshop}

\usepackage[utf8]{inputenc} 
\usepackage[T1]{fontenc}    
\usepackage{hyperref}       
\usepackage{url}            
\usepackage{booktabs}       
\usepackage{amsfonts}       
\usepackage{nicefrac}       
\usepackage{microtype}      
\usepackage{xcolor}         
\usepackage{graphicx}
\usepackage{subcaption}
\usepackage{tcolorbox}
\usepackage{longtable}
\usepackage{multirow}
\usepackage{wrapfig}

\title{Towards Time-Series Reasoning with LLMs}

\author{
  Winnie Chow\textsuperscript{1}\thanks{Work done during an internship at Apple.},
  Lauren Gardiner\textsuperscript{2},
  Haraldur T. Hallgr\'imsson\textsuperscript{2}, \and
  \textbf{Maxwell A. Xu\textsuperscript{3}\footnotemark[1],
  Shirley You Ren\textsuperscript{2}}
  \and
  \textsuperscript{1 }Stanford University, \textsuperscript{2 }Apple, \textsuperscript{3 }University of Illinois at Urbana-Champaign \and
  \texttt{wychow@stanford.edu}
}

\begin{document}

\maketitle

\begin{abstract}
   Multi-modal large language models (MLLMs) have enabled numerous advances in understanding and reasoning in domains like vision, but we have not yet seen this broad success for time-series. Although prior works on time-series MLLMs have shown promising performance in time-series forecasting, very few works show how an LLM could be used for time-series reasoning in natural language. We propose a novel multi-modal time-series LLM approach that learns generalizable information across various domains with powerful zero-shot performance. First, we train a lightweight time-series encoder on top of an LLM to directly extract time-series information. Then, we fine-tune our model with chain-of-thought augmented time-series tasks to encourage the model to generate reasoning paths. We show that our model learns a latent representation that reflects specific time-series features (e.g. slope, frequency), as well as outperforming GPT-4o on a set of zero-shot reasoning tasks on a variety of domains.

\end{abstract}

\section{Introduction}
\label{sec:introduction}

Multi-modal large language models (MLLMs) have enabled numerous advances in reasoning in domains such as vision \citep{liu2024visual, dai2023instructblipgeneralpurposevisionlanguagemodels}. They not only demonstrate the ability to act as controllers and decision-makers, but also show generalization ability to unseen tasks \citep{yin2024surveymultimodallargelanguage}. However, we have not yet seen this broad success for time-series. This modality is crucial for applications such as health coaching, financial investing, and environmental monitoring, where precise, human-interpretable temporal insights are essential. In particular, a time-series MLLM could be used to analyze and draw insights about a given time-series with natural language that follows human logic. 


Despite this need, enabling models to output natural language that reasons about a time series remains relatively unexplored, and there exists some skepticism of whether LLMs can zero-shot reason about time series \cite{merrill2024language}. In order to address this gap, we argue there are three essential steps to achieving time-series reasoning: \textbf{(1) Perception} -- understanding and identifying key characteristics in time-series data. \textbf{(2) Contextualization} -- extracting task-relevant features based on provided textual context. \textbf{(3) Deductive reasoning} -- drawing a conclusion based on the observations. 

We hypothesize that existing time-series MLLMs suffer from a perception bottleneck. In other words, the specific ways in which time-series are represented may limit how an LLM can reason about it. Specifically, many time-series MLLMs convert the time-series data into text tokens \citep{gruver2024largelanguagemodelszeroshot, liu2023largelanguagemodelsfewshot}, which could cause a loss in their ability to recognize temporal patterns. Additionally, reasoning capabilities might be emergent abilities observed only in larger models \citep{huang2022towards}, which implies that smaller models inherently lack the capability of contextualizing the time series and applying deductive reasoning.

In order to effectively recognize temporal patterns to address (1), we first train a lightweight time-series encoder on top of an LLM. We then show that the latent LLM representations encode various time-series features such as frequency and magnitude, even for out-of-distribution data. Next, we train with supervised fine-tuning on tasks augmented with chain-of-thought reasoning, which promotes the learning of reasoning processes described in (2) and (3), even if a small LLM is used. By addressing all three steps, our model with a smaller 7B-parameter LLM surpasses GPT-4o \citep{GPT-4o} on a set of zero-shot reasoning tasks on a variety of time-series domains.

\section{Related work}



\subsection{Time-series forecasting with LLMs}
The integration of MLLMs with time-series data has primarily been focused on traditional time-series analysis tasks, particularly forecasting \citep{zhang2024largelanguagemodelstime, chang2024llm4tsaligningpretrainedllms, zhou2023fitsallpowergeneraltime, jin2024timellmtimeseriesforecasting}. In these approaches, pretrained LLMs are typically used as the backbone, with special modules attached to capture properties of the time-series and align them with the LLM. However, these methods typically do not retain the language modeling head of the LLM and are not designed to output text. 


\subsection{Time-series question answering with LLMs}

A line of research has investigated the few-shot and zero-shot capabilities of LLMs in time-series question answering, typically by representing time series as text \cite{liu2023largelanguagemodelsfewshot}. 
Such modality conversion could result in a loss of critical information. Supporting this hypothesis, \cite{belyaeva2023multimodalllmshealthgrounded} showed that using a pretrained encoder to encode spirometry data directly achieves better downstream performance compared to serializing time-series as text. Also, \citet{merrill2024language} developed a dataset to assess general time-series reasoning and showed that none of their benchmarks significantly outperformed random chance.


\section{Methodology}
\subsection{Architecture}
\begin{wrapfigure}{r}{0.5\textwidth}
    \centering
    \vspace{-2.3cm}
    \includegraphics[width=\linewidth]{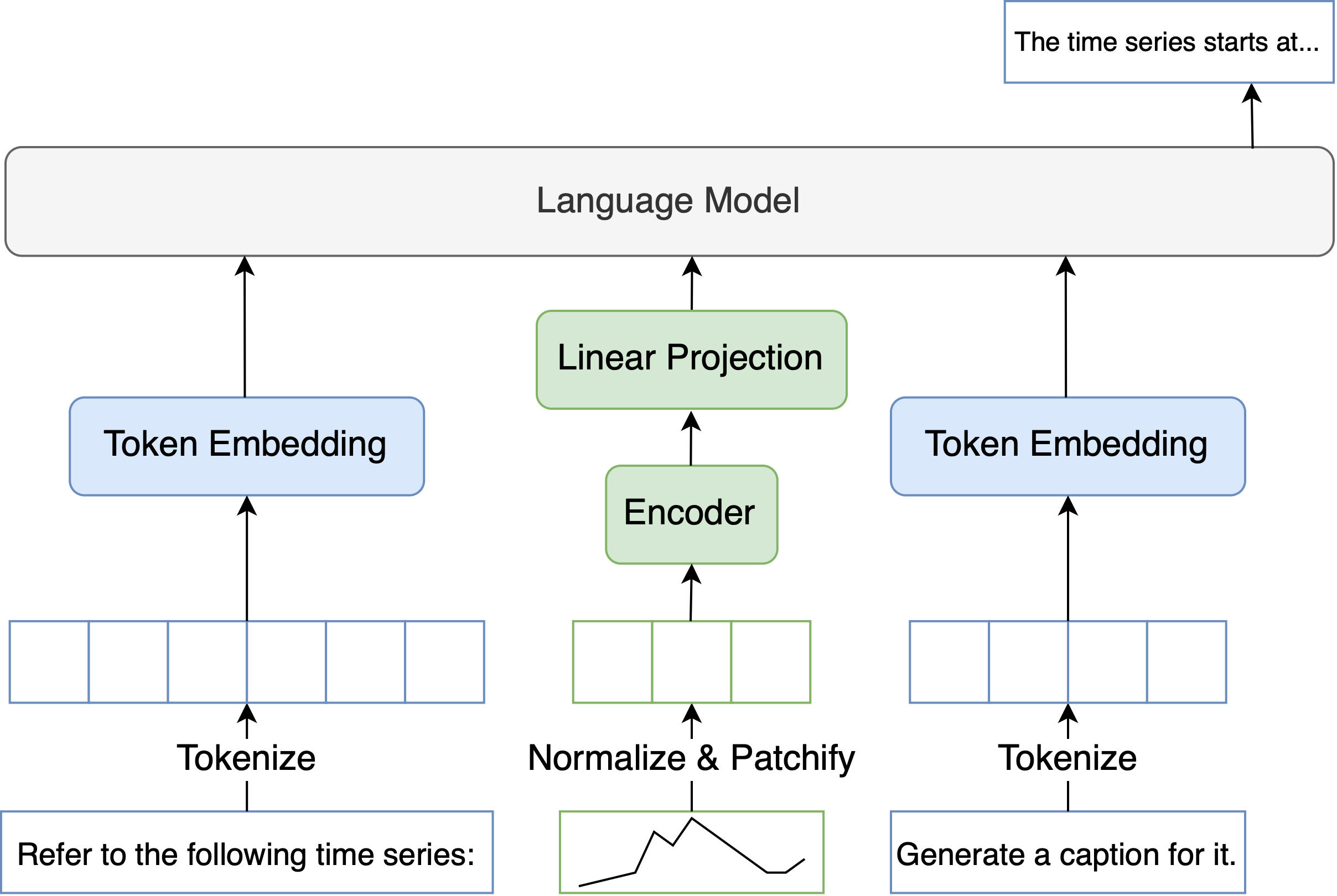}
    \caption{Proposed model architecture. Textual inputs are handled regularly, while time-series inputs are first normalized and divided to patch, pass through an encoder, then projected to the same dimension as the LLM's word embedding space.}
    \label{fig:architecture}
\vspace{-0.5cm}
\end{wrapfigure}
We first divide the input time-series into fixed, non-overlapping patches \citep{nie2022time} and feed these patches into a multi-head self-attention encoder. This is then followed by a linear layer which projects the features to match the dimensionality of the LLM's word embedding. Note that the input time-series is normalized, and its mean and variance are prepended in front of the time series tokens as text.

The text embedding of our prompts is concatenated with the time-series embedding and fed into the LLM. We can then generate text in an autoregressive fashion. For our experiments, we use a pre-trained Mistral-7B \citep{jiang2023mistral} as the LLM backbone. 


This design offers significant flexibility in handling varying input formats. First, it allows freely interleaving time-series and text. For example, we can put the text embedding first then concatenate the time-series afterward or vice versa. It can also handle multiple different time-series from different domains and multivariate time-series by treating each dimension independently. For example, given a 3-channel accelerometry signal, with each channel corresponding to the x,y,z axis respectively, we can simply feed in each axis sequentially while prepending a text embedding that describes which axis the signal is from. Finally, the self-attention mechanism imposes no restrictions on the length of the time-series, provided it remains within the LLM's context length and memory constraints.

\subsection{Training}
We adopt a two-stage training approach outlined below. We make use of a range of public time-series datasets and also generated synthetic time-series, which are detailed in Appendix \ref{appendix:datasets}. Specific language modeling tasks are generated from a mix of pre-defined templates or via GPT-4o. Each task's instruction includes 10-20 paraphrases of it in order to increase linguistic diversity. Refer to Appendix  \ref{appendix:training-details} for specific details on our training not captured here.

\begin{wrapfigure}{r}{0.4\textwidth}
  \centering
  \vspace{-0.1cm}
  \includegraphics[width=0.4\textwidth]{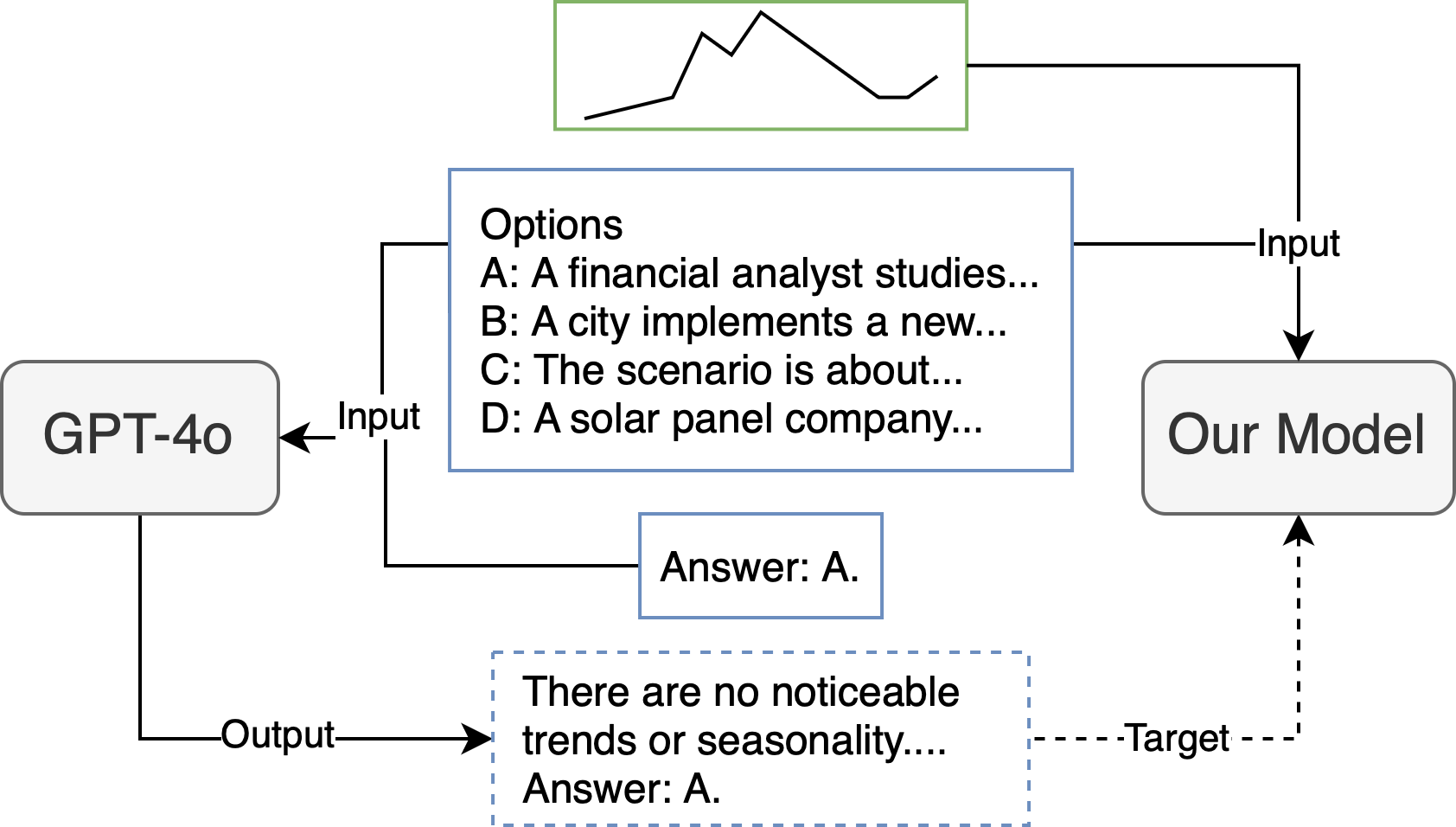}
  \caption{CoT augmentation for the etiological reasoning dataset. GPT-4o receives the four options, the answer label, and is prompted to generate a rationale for the correct option. }
  \label{fig:rationale_generation}
  \vspace{-0.5cm}
\end{wrapfigure}
\textbf{Stage 1: Encoder warm-up}. 
We train the encoder and projection layer from scratch for next token prediction while keeping the LLM frozen. However, achieving convergence is challenging when the model is trained from scratch \citep{tennenholtz2023demystifying}. Therefore, we adopt a curriculum learning approach. First, we train on a simple multiple-choice question-answering task on synthetic time-series. Then, we transition to a captioning task on synthetic data, and finally a captioning task on real data.


\textbf{Stage 2: Supervised fine-tuning on reasoning tasks}.
In the next stage, we fine-tune the encoder, projection layer, and the LLM end-to-end using LoRA \cite{hu2021loralowrankadaptationlarge} on a mixture of downstream tasks, most of which augmented with GPT-generated chain-of-thought (CoT) text \citep{wei2022chain} (Figure \ref{fig:rationale_generation}). CoT allows the model to learn to reason through a logical flow of natural language and to exploit the encoded time-series features.



\section{Experiments}
\label{sec:encoder}
\textbf{(1) Perception: Does the encoder help the LLM better perceive time-series data?}

The first step is to verify whether the model is able to encode general time-series features after training stage 1. To achieve this, we assess the usefulness of model-generated captions on downstream tasks by feeding the caption (instead of raw time series) to a strong reasoner (GPT-4o). Example captions can be found in Appendix \ref{appendix:generation-example}.
We evaluate our model with etiological reasoning, proposed by \cite{merrill2024language}, which consists of multiple choice questions of the most likely process to have generated a given time-series. Since this task requires matching specific time-series features to characteristics of the domain as well as specific events affecting the time-series, good performance on this task would imply that the captions capture useful temporal information.


Table \ref{tab:caption_performance} shows that using captions generated by our Mistral-7B-based model after training stage 1, despite under-performing GPT-4o captions, shows a significant improvement from using captions generated by text-only Mistral-7B. This supports our hypothesis that encoding time-series directly is more efficient than converting it to text or an image. A full analysis of disentangling the effects of time-series representation and the reasoner is in Appendix \ref{appendix:reasoning-bottleneck}. 

\begin{table}[]
  \caption{Accuracy of GPT-4o reasoner on etiological reasoning when taking in captions generated by the proposed model. For comparison,
  we also generate captions by making GPT-4o take in a plot and Mistral-7B take in time-series as text. Caption generated by our model gives a significant performance boost compared to converting time series to text.} \
  \label{tab:caption_performance}
  \centering
  \resizebox{0.45\textwidth}{!}{%
  \begin{tabular}{c|cc}
    \toprule
     Our Model & GPT-4o (plot) & Mistral-7B (text)\\
    \midrule
    0.387 & 0.455 & 0.272\\
    \bottomrule
  \end{tabular}
  }
\vspace{-0.5cm}
\end{table}

\begin{figure}[ht!]
    \centering
    \vspace{-0.5cm}
    \includegraphics[width=0.65\linewidth]{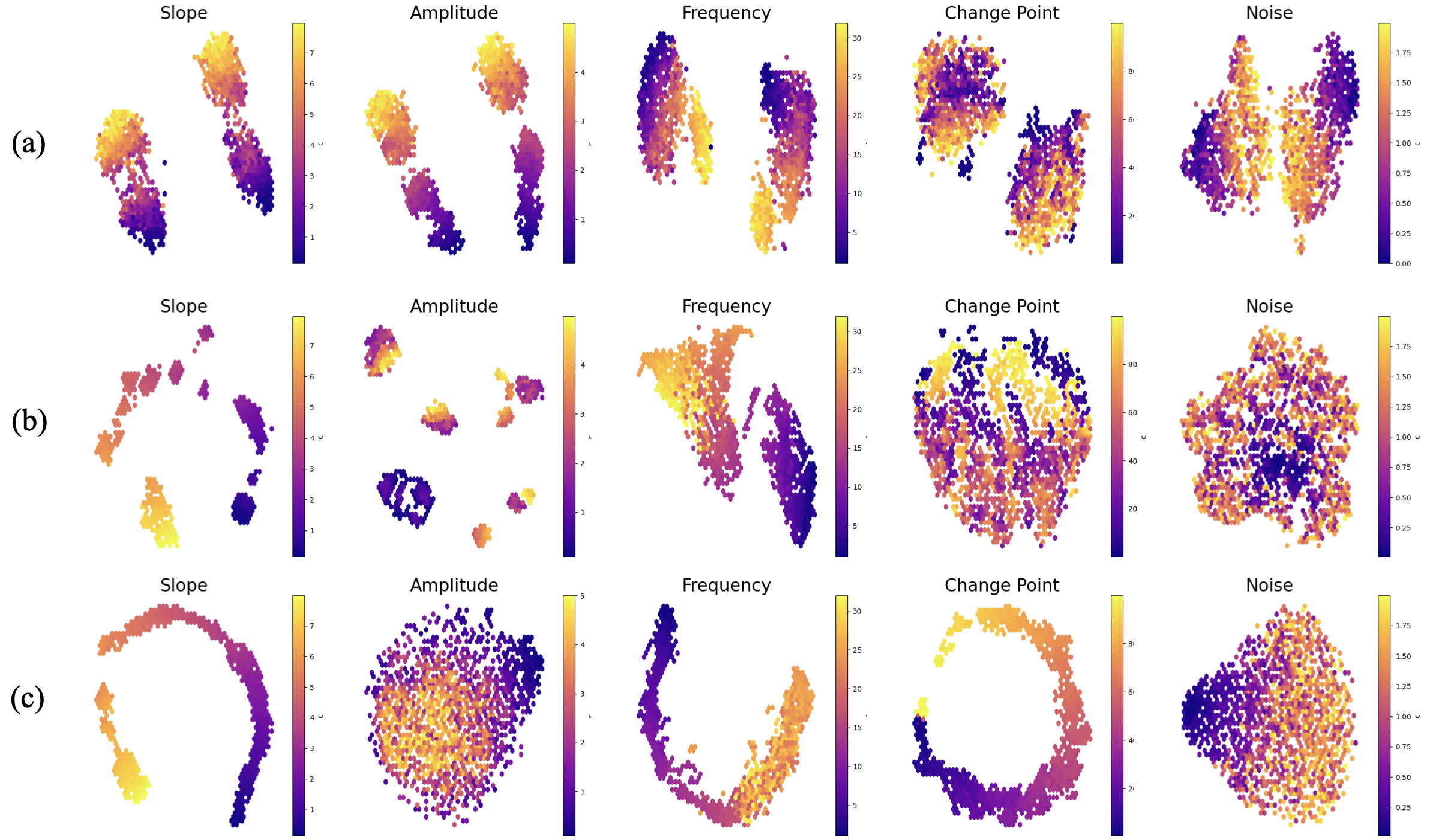}
    \caption{t-SNE visualization of the hidden states of the LLM with a synthetic sine wave input. The synthetic sine wave is continuously varied along one of the five given characteristics for each column. Each row represents a different method in which the time-series is being used with the LLM: (a) converting the time-series into text for Mistral-7B input, (b) using an \textbf{un}trained time-series encoder with our fused LLM approach, (c) using a trained encoder with our approach. The color of each point represents the given value of $c$, where $c$ is the value of one of the five time-series characteristics. Continuous changes in a given parameter $c$ correspond to continuity within the latent space, suggesting that the encoder has aligned time-series features to the LLM.     
    }
    \label{fig:hidden-toy}
    \vspace{-0.5cm}
\end{figure}

Next, in Figure. \ref{fig:hidden-toy}, we visualize the last hidden states of the LLM to understand how specific time-series characteristics are easily accessible to the LLM, promoting perception. Appendix \ref{appnedix:encoder-visualization} details the dataset used and additional qualitative analysis.





\textbf{(2, 3) Contextualization and Deductive Reasoning: Evaluating reasoning capabilities through zero-shot classification}

In order for a model to achieve nontrivial zero-shot performance for a given task, this model could first contextualize the time-series into its most relevant features, then apply deductive reasoning to answer the task. Therefore, to evaluate the model's reasoning capabilities on unseen tasks, we define zero-shot classification tasks on selected datasets from the UCR Classification Archive \citep{UCRArchive2018}. We select a subset of datasets which we deem most suitable for zero-shot classification. Refer to Appendix \ref{sec:zero-shot dataset} for details and rationale regarding the selection process, time-series samples from the dataset, and an example of the model generated reasoning path on a task.

We note that these tasks are demanding even after the selection -- some of the datasets are normalized, and most contain dataset specific features that we would not expect the model to know merely by reasoning. For example, the DodgerLoopGame dataset contains traffic data collected near a stadium either from a game day or a normal day. Data recorded on a game day usually have peak values greater than 50, while values are usually under 50 on a normal day. A model that has not seen these data would not be able to deduce the absolute value (50) of such a threshold.

\begin{table}[ht]
    \centering
    \vspace{-0.4cm}
    \caption{Zero-shot classification compared to GPT-4o with time-series input as text or plot}
    \label{tab:zero-shot-performance}
    \resizebox{.8\textwidth}{!}{%
    \begin{tabular}{l|cc|cc|cc}
    \toprule
     & \multicolumn{2}{c|}{Our Model} & \multicolumn{2}{c|}{GPT-4o (text)} & \multicolumn{2}{c}{GPT-4o (plot)} \\
     & Accuracy & F1 & Accuracy & F1 & Accuracy & F1 \\
    \midrule
    Chinatown & \textbf{0.698} & \textbf{0.602} & 0.347 & 0.233 & 0.287 & 0.133 \\
    Computers & \textbf{0.581} & \textbf{0.552} & 0.545 & 0.539 & 0.564 & 0.548 \\
    DodgerLoopGame & 0.476 & 0.347 & \textbf{0.515} & \textbf{0.357} & 0.495 & 0.000 \\
    DodgerLoopWeekend & \textbf{0.654} & 0.395 & 0.554 & 0.498 & 0.594 & \textbf{0.527} \\
    ECG200 & 0.435 & 0.399 & 0.424 & 0.418 & \textbf{0.566} & \textbf{0.543} \\
    GunPointAgeSpan & \textbf{0.587} & 0.587 & 0.525 & 0.362 & 0.505 & 0.000 \\
    HouseTwenty & 0.514 & \textbf{0.448} & \textbf{0.525} & 0.261 & 0.416 & 0.196 \\
    ItalyPowerDemand & \textbf{0.701} & \textbf{0.682} & 0.564 & 0.552 & 0.505 & 0.487 \\
    MoteStrain & 0.473 & 0.459 & 0.515 & \textbf{0.491} & \textbf{0.525} & 0.460 \\
    PowerCons & \textbf{0.643} & \textbf{0.626} & 0.495 & 0.377 & 0.485 & 0.327 \\
    SonyAIBORobotSurface1 & \textbf{0.642} & \textbf{0.575} & 0.436 & 0.413 & 0.416 & 0.416 \\
    \bottomrule
    \end{tabular}
    }
\end{table}


From Table \ref{tab:zero-shot-performance}, we see that the model performs significantly above chance on a majority of datasets. As our model has not been trained on these datasets or tasks, these results show evidence that the model can generalize to unseen tasks via CoT reasoning. In contrast, GPT-4o is consistently close to chance, which is most likely because of the lack of a time-series specific encoder to properly perceive and integrate time-series features.


\section{Conclusion}
In this work, we first decompose the capabilities needed for an LLM to reason about time-series data, then posit that the representation of the time-series data will heavily impact performance. As such, we utilize a lightweight but flexible time-series encoder that greatly improves the LLM's ability the perceive the data. Combined with supervised fine-tuning on chain-of-thought augmented data, our approach demonstrates nontrivial generalization ability to unseen tasks, even surpassing GPT-4o. Our work has helped show how multi-modal language models can be used to generate human-interpretable text about a time-series that follows human logic, and we hope that our work will help unlock new applications at the intersection of time-series analysis and complex decision-making.


\bibliographystyle{unsrtnat}
\bibliography{ref}

\appendix

\section{Acknowledgements}
We would like to thank Vincent Chan, Feng Zhu, Fuli Wang from Apple, and Ryan Chan from University of Pennsylvania for their valuable discussion or feedback.

\section{Case study on the reasoning bottleneck}
\label{appendix:reasoning-bottleneck}

In this section we focus on etiological reasoning, one of the benchmarks proposed by \citet{merrill2024language}. In each question, the model is given a time-series (without any metadata such as unit of measurement) with four scenarios and has to select the one that most likely produced the time-series. It is a valuable benchmark to access models' ability to integrate characteristics from time-series in conducting multi-step reasoning: 1) hypothesizing features that would be present in the time-series for each option 2) comparing and contrasting to select the option that best matches with the given time-series. 

To disentangle the effects of time-series representation and reasoning capabilities, we evaluate different variants of the GPT family and Mistral-7B, an open-source model, by varying the time-series representation scheme. We consider the following methods: \textbf{Comma-separated}: the time-series is represented as a comma-separated sequence of numbers, without any processing. \textbf{Tokenization}, adopted by \citet{merrill2024language}, encodes time-series following \citet{gruver2024largelanguagemodelszeroshot}'s tokenization methodology. \textbf{Plot} with different figure sizes are considered, with (3, 0.4) intended to replicate the figure size used in \citet{merrill2024language}. \textbf{Caption} uses natural language description of the time-series generated by GPT-4o, aiming to bridge any potential modality gap.

\begin{table}[ht!]
  \caption{Models' performance on etiological reasoning with different input types.}
  \label{understanding_results}
  \centering
  \begin{tabular}{lllllll}
    \toprule
    \cmidrule(r){1-2}
    Input & GPT-4o & GPT-4 & GPT-3.5 & Mistral-7B & Llama-7B & Llama-13B\\
    \midrule
    Comma separated & 0.441 & 0.394 & 0.253 & 0.238 & - & -\\
    Tokenization & 0.322 & 0.340 & 0.218 & 0.261 & 0.273 & 0.278 \\
    \midrule
    Plot (3, 0.4) & 0.410 & 0.32 & - & - & - & - \\
    Plot (3, 3) & 0.56 & 0.38 &  - & - & - & -\\
    Plot (10, 3) & 0.609 & 0.42 & - & - & - & - \\
    \midrule
    Caption & 0.455 & 0.460  & 0.255 & 0.242 & - & -\\
    
    
    \midrule
    \multicolumn{7}{l}
    {Random baseline: 0.25 \quad Human: 0.661} \\
    \bottomrule
  \end{tabular}
\end{table}


\textbf{GPT-4o shows non-trivial ability to solve the task,} with performance increasing with higher resolution plotting. This is consistent with observations in the literature that reasoning is an emergent ability. \citep{yin2024surveymultimodallargelanguage}. Due to its promising performance and the lack of a ground truth time-series caption, we generate detailed captions for each time-series using GPT-4o to serve as a proxy to a ground truth label.

\textbf{Time-series representations have a significant impact on performance.} For example, performance almost doubled when plotting the time-series compared to tokenization. It is also observed that the higher the plot resolution, the better the performance for the two models with vision capabilities.

\textbf{Models' struggle with reasoning is not solely caused by modality misalignment.} Feeding the proxy ground truth caption in natural language, though potentially causing loss of information, should eliminate any modality gap, thus eliminating the possibility of models failing due to not being able to perceive the time-series. GPT-4 shows a performance gain taking in the caption compared to text, suggesting that it does encode information useful for the task. However, no other models display a similar performance gain, reflecting that the bottleneck also lies in the model's reasoning capabilities.

We observe that representing time-series as text is inefficient for helping the model perceive the time-series. Plotting, despite relatively better performance, is very sensitive to hyperparameters such as the figure size. A 10 x 3 figure is required for GPT-4o to get close to human performance. This motivates designing architectures that allow the model to perceive the time-series in a more native manner instead of relying on transforming into a different modality.

The main takeaways are that 1) reasoning capabilities are only evident in bigger models 2) time-series bottleneck is faced by all. Therefore, to develop a model that is able to reason about time-series, we would need to tackle both of these challenges.

\section{Datasets}
\label{appendix:datasets}

Annotated time series data generally remains quite sparse due to the high cost of annotation, normally requiring domain-level expertise (e.g. identifying motifs in electrocardiogram signals \citep{xu2024rebar}). Therefore, in addition to real datasets, we generate synthetic datasets, from which we can easily generate annotations.

 We follow the train/validation/test split specified by creators of the dataset whenever available.

\textbf{PAMAP2} \citep{reiss2012introducing}. Recorded from 18 activities performed by 9 subjects, wearing 3 IMUs and a HR-monitor. Only accelerometer data from the device on users' dominant hand is considered. Following the original paper, subject 5 and 6 are used as test subjects. We split the time-series with non-overlapping two-second windows and only consider the 12 main activities.

\textbf{Sleep} \citep{867928}. 1-lead electroencephalography (EEG) signals sampled at 100 Hz from 153 whole-night sleeping EEG recordings take from physionet. The series were segmented into non overlapping sub series, each falling into one of the five sleeping patterns/stages: wake, non rapid eye movement type 1, non rapid eye movement type 2, non rapid eye movement type 3, and rapid eye movement.

\textbf{M4} \cite{Makridakis2020TheMC}. Consists of time-series of yearly, quarterly, monthly and other (weekly, daily and hourly) data, totaling 100,000 time-series used for the fourth edition of the Makridakis forecasting Competition.

\textbf{UCR Archive} \citep{UCRArchive2018}. A collection of 128 datasets covering time-series classification tasks in various domains. Some datasets are not used in training and reserved for evaluation.

\textbf{Etiological Reasoning} \citep{merrill2024language}. Synthetic dataset consisting of time-series generated with code, along with metadata used for generation such as the scenario and characteristics of the series.





\textbf{Trend}. Synthetic dataset of time-series with different kinds of trends, generated by combining various monotonic functions.
    
\textbf{Pattern}. Synthetic dataset generated from adding a fixed set of features to a base series, namely trend, seasonality, outliers, and level shift. Captions are generated from combining templates for the four features in a random order.



\section{Training details and tasks}

\label{appendix:tasks}
Below is a list of task types considered in this work:
 
\textbf{Captioning}. Given a time-series without prescribed context, generate a description of its features and value changes over time. We consider two variants: short captions (1-2 sentences), and long captions (a paragraph). Refer to Figure \ref{fig:m4-long} for an example task defined on the M4 dataset.

\begin{figure}[ht!]
    \begin{tcolorbox}[title=TrendMCQ]
    \textbf{Input:}\\
    \begin{center}
        \includegraphics[width=0.7\textwidth]{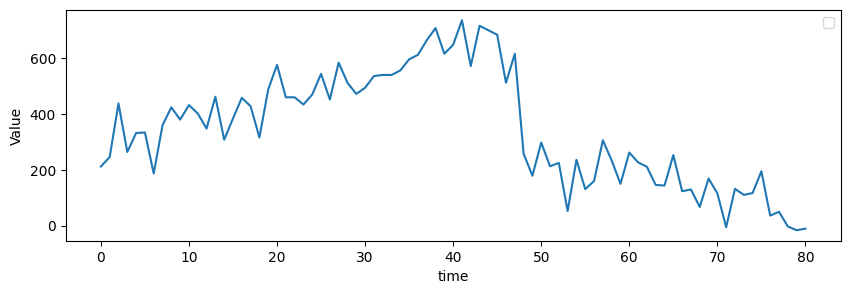} 
    \end{center}
    How would you describe the trend in this time-series: (mean: 346.97, standard deviation: 202.85) <TS>?\\
    A: increasing.\\
    B: increasing then decreasing.\\
    C: decreasing then increasing.\\
    D: stationary.\\
    Respond with one letter: "A", "B", "C", or "D".
    \\
    
    \textbf{Output:} B.
    \end{tcolorbox}
    
    \begin{tcolorbox}[title=PatternMCQ]
    \textbf{Input:}\\
    \begin{center}
        \includegraphics[width=0.7\textwidth]{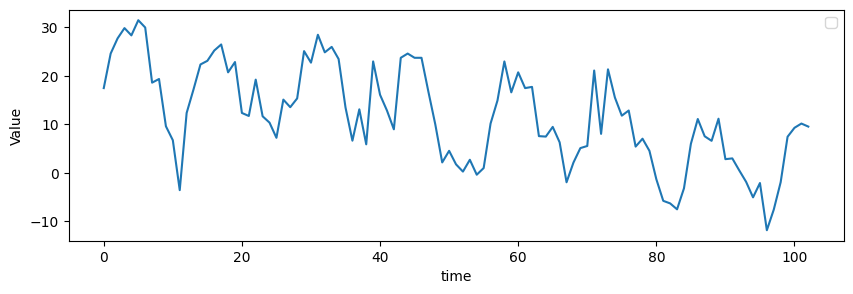} 
    \end{center}
    Data: (mean: 11.80, standard deviation: 10.06) <TS>\\
    Which of the following descriptions aligns best with the time-series shown above?\\
    A: No upward or downward trend is observed. The series is free from unusual values. A periodic pattern occurs every 18 years. The time-series shows a level shift later in the series.\\
    B: No noticeable level shifts are detected. The series shows a downward trend. A periodic pattern occurs every 14 seconds. No anomalous points are detected.\\
    C: There is a distinct level shift in the middle of the series. The data contains 4 significant deviations. The time-series follows a every 32 minutes seasonal cycle. The sequence is overall trending upward.\\
    D: The data contains 12 significant deviations. A periodic pattern occurs every 22 minutes. The trend of the series is decreasing. The series does not show any abrupt changes.\\
    Answer only with the letter corresponding to the correct choice.\\
    
    \textbf{Output:} B.
    \end{tcolorbox}
\caption{Data examples, where <TS> is replaced by tokens from time-series.}
\label{fig:mcq_examples}
\end{figure}

\textbf{Question Answering (QA)}. Given one or more time-series, answer questions about their features. This could be accompanied with context. Refer to Figure \ref{fig:mcq_examples} for two QA tasks formatted as multiple choice questions.

\textbf{Classification}. Given a time-series along with its context and possible classes in natural language, identify the class the time-series most likely corresponds to.

\textbf{Etiological Reasoning}. Given a time-series, the model hypothesize about the scenario from which it was generated.

Each task can either be in the form of multiple choice questions (MCQs), choosing among a list of options, or free-form response. For instructions that do not expect free-form response, the model is explicitly told the output space. For example, for a Yes/No question, the instruction ends with a phrase like "Answer only with "Yes" or "No". This helps alleviate the potential issue of the model learning representations which help condition the response in terms of the response format \citep{liu2024improved}. 

\label{appendix:training-details}
\begin{table}[ht!]
  \caption{Tasks used for pretraining and finetuning.}
  \label{tab:training_tasks}
  \centering
  \begin{tabular}{lcccc}
    \toprule
     Task & Underlying Dataset & Size (Train/Val/Test) & Proportion \\
    \midrule
    \textbf{Stage 1} & & & & \\
    MCQ & Trend & 22k/1k/1k & 1\\
    \cmidrule(r){1-4}
    Captioning & Pattern & 36k/1k/1k & 1\\
    \cmidrule(r){1-4}
    Captioning (short) & M4 & 3k/1k/1k & 0.15\\
    Captioning (long) & M4 & 3k/1k/1k & 0.35\\
    Captioning (short) & UCR & 3k/1k/1k & 0.15\\
    Captioning (long) & UCR & 3k/1k/1k & 0.35\\
    \cmidrule(r){1-4}
    
    \textbf{Stage 2} & & & \\
    Etiological reasoning & Etiological reasoning & 6k/500/1k & 0.60\\
    Classification & PAMAP2 & 10k/1k/1k & 0.15\\
    Classification & Sleep & 10k/1k/1k & 0.15\\
    Captioning (long) & M4 & 3k/1k/1k & 0.05\\
    Captioning (long) & UCR & 3k/1k/1k & 0.05\\
    
    \bottomrule
  \end{tabular}
\end{table}

Table \ref{tab:training_tasks} shows the list of tasks used in different stages of training.


\begin{table}[ht!]
  \caption{Hyperparameters used in each training stage.}
  \label{tab:hyperparameters}
  \centering
  \begin{tabular}{c|cc}
    \toprule
     Hyperparameter & Stage 1 & Stage 2 \\
    \midrule
    Steps & 3000/3000/3000 & 5000\\
    Batch size & 64 & 64\\
    Learning rate & 8e-4 & 2e-5\\
    Warmup ratio & 3\% & 3\% \\
    LR schedule & linear & linear \\
    
    \bottomrule
  \end{tabular}
\end{table}

A patch size of 4 is used. We use a multi-head attention layer with 12 heads and dimension=600.
Table \ref{tab:hyperparameters} gives the hyperparameters used. For all stages, we train the model on 8 $\times$ A100 GPUs.

\section{What did the encoder learn?}
\label{appnedix:encoder-visualization}

\begin{figure}[ht!]
    \includegraphics[width=0.5\linewidth]{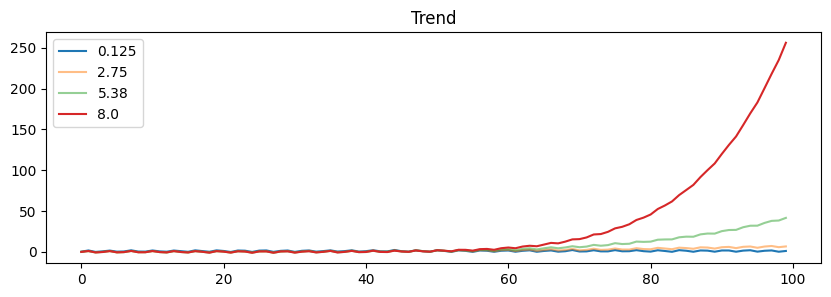}
    \includegraphics[width=0.5\linewidth]{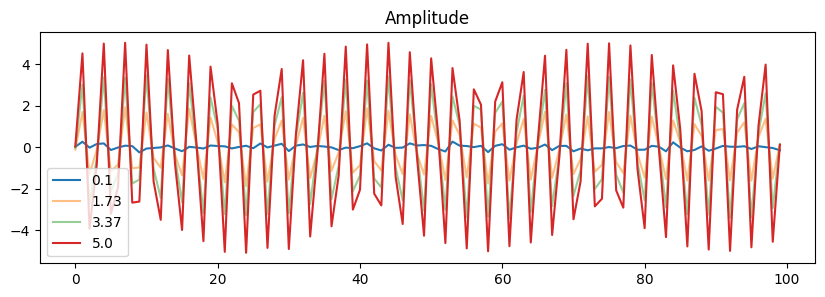}
    \includegraphics[width=0.5\linewidth]{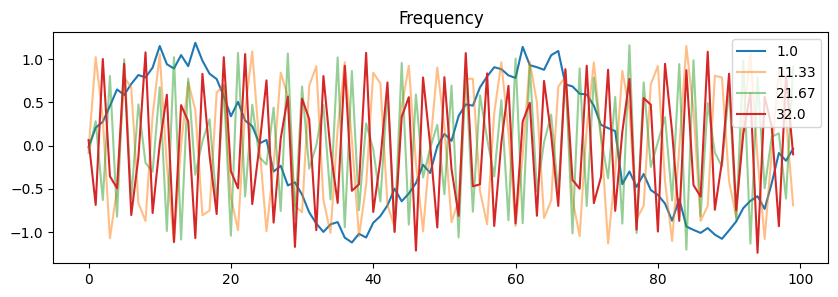}
    \includegraphics[width=0.5\linewidth]{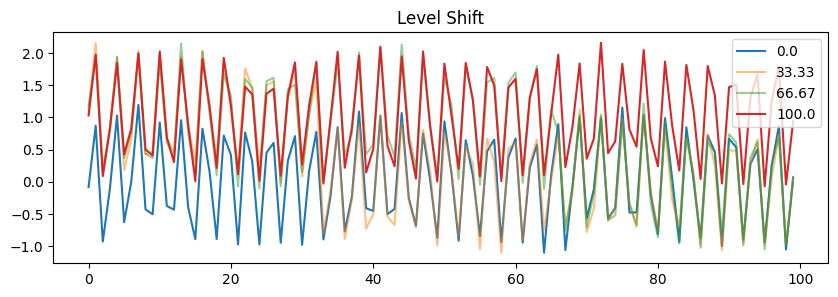}
    \includegraphics[width=0.5\linewidth]{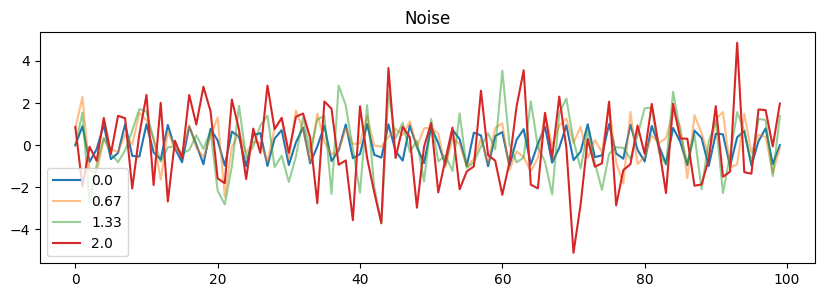}
\caption{Toy datasets used to generate the visualization in Figure \ref{fig:hidden-toy}.}
\label{fig:toy_datasets}
\end{figure}

\begin{figure}[ht!]
    \centering
    \begin{tabular}{cc}
        \begin{subfigure}[b]{0.5\textwidth}
            \centering
            \includegraphics[width=\textwidth]{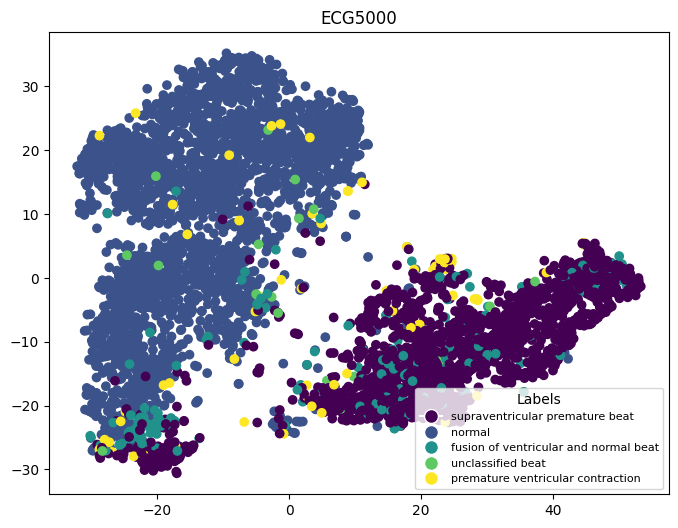}
        \end{subfigure} &
        \begin{subfigure}[b]{0.5\textwidth}
            \centering
            \includegraphics[width=\textwidth]{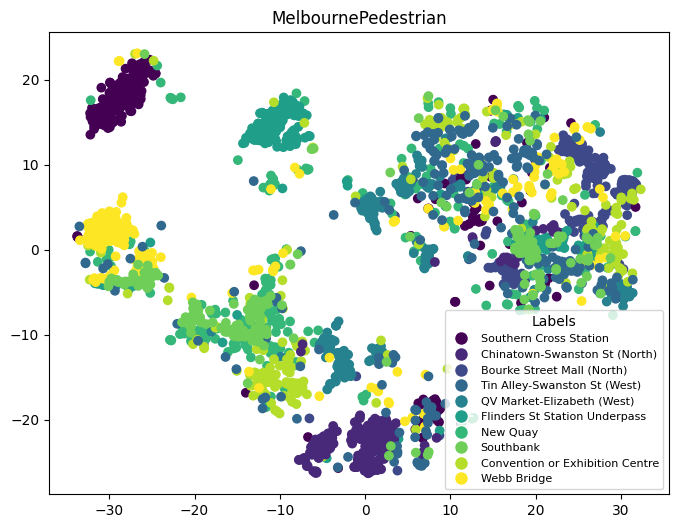}
        \end{subfigure} \\
    \end{tabular}
    \caption{t-SNE visualization of hidden states of the LLM on UCR classification archive datasets with large test sets. Note that the model has never seen these datasets during training.}
    \label{fig:vis-ucr}

\end{figure}

The toy datasets used to generate the visualization in \ref{sec:encoder} are shown in Figure \ref{fig:toy_datasets}. Figure \ref{fig:vis-ucr} visualizes how time-series of different classes are clustered for two UCR datasets.

\section{UCR Classification Archive dataset selection protocol}
\label{sec:zero-shot dataset}
\begin{enumerate}
    \item Select only binary classification tasks due to the challenging nature of zero-shot classification.
    \item Select only datasets with a test size of at least 100.
    \item Exclude pseudo-time-series, such as those derived from outline of images. The rationale behind is that they don't have a natural interpretation.
    \item Select only datasets whose description provides enough context. For example, we exclude datasets with no named labels.
    \item Exclude classification tasks that are heavily dependent on the training examples and not solvable by commonsense reasoning, such as classifying spectrographs.
\end{enumerate}

\begin{figure}[ht!]
    \begin{minipage}{0.5\textwidth}
        \centering
        \includegraphics[width=\textwidth]{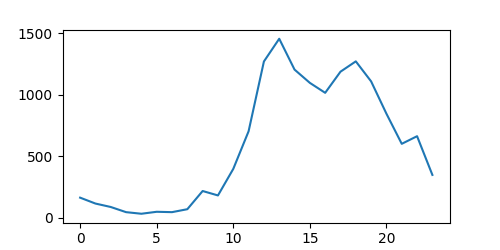}
        \\ 
        Weekday
    \end{minipage}%
    \begin{minipage}{0.5\textwidth}
        \centering
        \includegraphics[width=\textwidth]{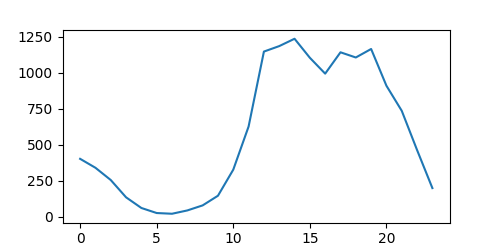}
        \\ 
        Weekend
    \end{minipage}
    \caption{Examples of each class from the Chinatown dataset, showcasing the subtle
    differences between them.}
    \label{fig:chinatown-plots}
    
\end{figure}

Following this procedure, we eliminate down to 11 datasets. Even with these steps, the tasks are still quite challenging as the difference between classes are often subtle. For example, in Figure \ref{fig:chinatown-plots}, a human can tell apart time-series of Chinatown pedestrian traffic on weekends compare to weekdays by hypothesizing that weekdays would show more distinct rush hour peaks.

We saw no significant difference in performance with and without chain-of-thought for GPT-4o. 

\section{Examples of model-generated outputs}
\label{appendix:generation-example}

\begin{figure}[ht!]
    \begin{tcolorbox}[title=M4-long (test set)]
    \textbf{Input:}\\
    \begin{center}
        \includegraphics[width=0.7\textwidth]{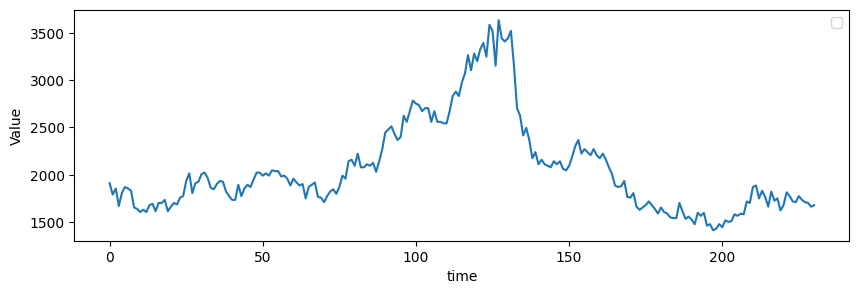} 
    \end{center}
    Generate a detailed caption for the following time-series data: <TS>.
    \\
    
    \textbf{GPT-generated output:}\\
    The time-series demonstrates a fluctuating pattern over a range of observations. Initially, the values hover around 1750 to 2000, showcasing minor oscillations. As the sequence progresses, a more pronounced increase becomes evident, with the values reaching a peak around 3500 near the 100th observation. Subsequently, a distinct decline occurs, lowering the values to approximately 2000. Periodic minor fluctuations continue to manifest past the 150th observation point, maintaining values in the range of 2000 to 2500. The end of the series reveals continued volatility with a slight downward trend, stabilizing around 2000 after some fluctuations.\\
    \textbf{Model-generated output:}\\
    The time-series data shows a general upward trend with fluctuations over a period of approximately \textcolor{red}{300 units}. Initially, the values \textcolor{blue}{start around 1500} and \textcolor{blue}{gradually increase}, reaching \textcolor{blue}{a peak of approximately 3500} around the \textcolor{red}{100th unit}. Following this peak, there is a \textcolor{blue}{significant decline}, with values dropping to around \textcolor{blue}{1500}. After this decline, the series exhibits a series of \textcolor{blue}{smaller fluctuations, oscillating between 1500 and 2500}. Towards the end, there is a \textcolor{blue}{slight upward trend}, with values hovering around \textcolor{blue}{2000}. The overall pattern suggests an initial rise, a sharp decline, and a period of relatively stable oscillations with minor fluctuations.
    \end{tcolorbox}
    \caption{Caption generated by the model for the test split of M4 after training stage 1, along with proxy ground truth generated by GPT-4o from a plot.}
    \label{fig:m4-long}
\end{figure}

\begin{figure}[ht!]
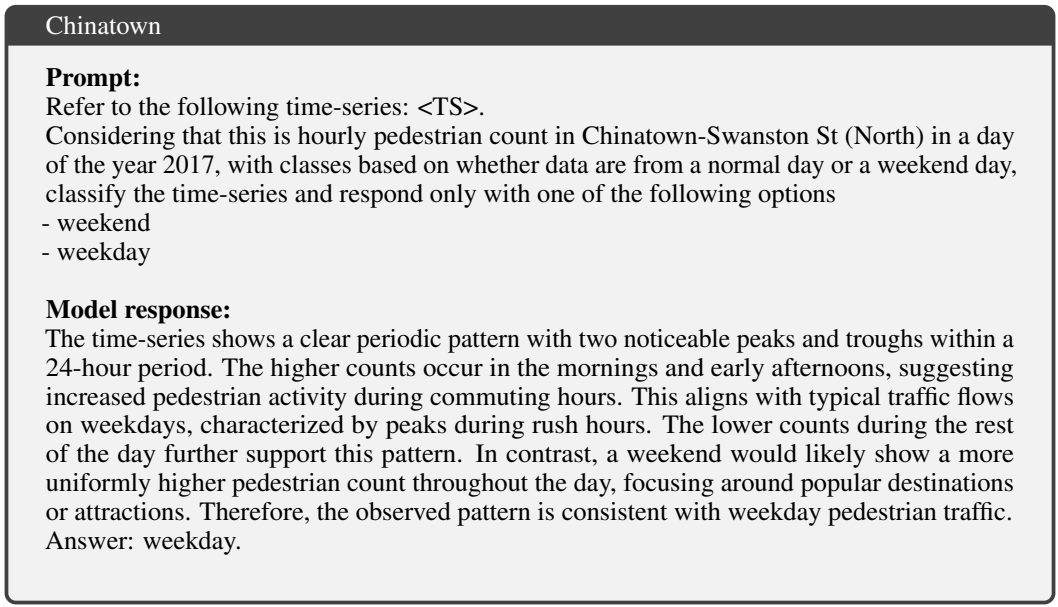

    \begin{tcolorbox}[title=Chinatown]

    \textbf{Prompt:}\\
    Refer to the following time-series: <TS>.\\
    Considering that this is hourly pedestrian count in Chinatown-Swanston St (North) in a day of the year 2017, with classes based on whether data are from a normal day or a weekend day, classify the time-series and respond only with one of the following options\\
    - weekend\\
    - weekday\\

    \textbf{Model response:}\\
    The time-series shows a clear periodic pattern with two noticeable peaks and troughs within a 24-hour period. The higher counts occur in the mornings and early afternoons, suggesting increased pedestrian activity during commuting hours. This aligns with typical traffic flows on weekdays, characterized by peaks during rush hours. The lower counts during the rest of the day further support this pattern. In contrast, a weekend would likely show a more uniformly higher pedestrian count throughout the day, focusing around popular destinations or attractions. Therefore, the observed pattern is consistent with weekday pedestrian traffic.\\
    Answer: weekday.
    \\
    \end{tcolorbox}
    \caption{Chain-of-thought classification on the Chinatown dataset from the UCR Archive.}
    \label{fig:zero-shot-example}
\end{figure}

Figures \ref{fig:m4-long} shows an example of a long caption generated by the model on unseen data. Figure \ref{fig:zero-shot-example} shows an example of the model's reasoning path on the Chinatown dataset.

\section{Limitations}

While our work has demonstrated potential in generating textual response based on both textual and time-series modalities, the generation of time-series data is left for future work. Accomplishing that would unlock notable capabilities such as contextualized forecasting. While this is technically possible with the current approach by asking the model to generate predictions in a textual format, we believe this is ineffective. Future work should also explore different architecture design choices such as the encoder architecture, better ways to incorporate mean and variance, etc. Finally, the model is currently trained on a limited set of tasks -- increasing the variety of tasks could potentially further improve the model's reasoning and generalization abilities.

\end{document}